\documentclass{article}
\usepackage{graphicx} 
\usepackage{amsmath, amssymb}
\usepackage{geometry}
\usepackage{algorithm,algpseudocode}
\usepackage{bbm}
\usepackage{xcolor}
\usepackage{graphicx}
\usepackage{float}
\usepackage[colorlinks=true, citecolor=green]{hyperref}

\title{Exploring Diffusion and Flow Matching Under Generator Matching}
\author{Zeeshan Patel, James DeLoye, Lance Mathias\\[1em]
    University of California, Berkeley}
\date{December 2024}

\begin{document}

\maketitle
\begin{abstract}
In this paper, we present a comprehensive theoretical comparison of diffusion and flow matching under the Generator Matching framework. Despite their apparent differences, both diffusion and flow matching can be viewed under the unified framework of Generator Matching. By recasting both diffusion and flow matching under the same generative Markov framework, we provide theoretical insights into why flow matching models can be more robust empirically and how novel model classes can be constructed by mixing deterministic and stochastic components. Our analysis offers a fresh perspective on the relationships between state-of-the-art generative modeling paradigms.
\end{abstract}
\section{Introduction}

Recent techniques in deep generative modeling have leveraged Markov generative processes to learn complex, high-dimensional probability distributions in a more structured and flexible manner~\cite{wu2019deepgenerativemarkovstate}. By integrating Markov chain methods with deep neural architectures, these approaches aim to exploit the representational power of deep networks while maintaining a tractable and theoretically grounded training procedure. In contrast to early generative models that relied heavily on direct maximum likelihood estimation or adversarial objectives, this class of methods employs iterative stochastic transformations—often expressed as Markovian updates—to gradually refine initial noise samples into samples drawn from the desired target distribution.

Diffusion and flow matching models represent two prominent classes of generative approaches that construct data samples through a sequence of continuous transformations. Diffusion models~\cite{ho2020denoising,pmlr-v37-sohl-dickstein15} introduce a forward-noising and reverse-denoising process, progressively refining a simple noise distribution into a complex target distribution by learning to undo incremental noise corruption at each step. Flow matching models~\cite{lipman2023flowmatchinggenerativemodeling,fairmatching,explicit-flow}, on the other hand, directly learn continuous-time transformations that morph a base distribution into the target distribution under a prescribed flow field. Both families benefit from well-defined likelihoods and stable training objectives, allowing for clearer theoretical insights, improved sample quality, and often more reliable convergence than prior approaches such as GANs~\cite{dhariwal2021diffusionmodelsbeatgans,goodfellow2014generativeadversarialnetworks}.

Generator Matching~\cite{genmatching} is a framework that unifies generative modeling with Markov processes on arbitrary state spaces. This framework allows combining different Markov processes in two ways: Markov superposition and creating multimodal generative models by combining unimodal generators. In this work, we aim to leverage the Generator Matching framework to provide a detailed theoretical comparison of diffusion and flow matching models. We show

we aim to provide an overview of Generator Matching, how it connects to diffusion and flow matching models, and how specific properties of some Markov generative process make them more robust than others.

\section{Connecting Diffusion and Flow Matching}
Diffusion and Flow Matching are two different generative Markov processes that have many similarities. In this section, we briefly define diffusion and flow matching models and provide some intuition on how diffusion and Gaussian flow matching models are theoretically equivalent.

\subsection{Diffusion Models}

Diffusion models learn to reverse a time-dependent destructive process that corrupts data $x$ by adding noise~\cite{ho2020denoising}. In practice, the learned component of a diffusion model is a neural network $f_{\theta}$ that tries to estimate the denoised data given the noisy sample $\mathbf{z_t} = \alpha_t \mathbf{x} + \sigma_t \boldsymbol{\epsilon}$ at time $t$, where typically $\boldsymbol{\epsilon} \sim \mathcal{N}(0, \mathbf{I})$ is Gaussian noise and $(\alpha_t, \sigma_t)$ define the noise schedule.

We can generate samples from diffusion models by reversing the forward corruption process. We initialize our sample as pure Gaussian noise, $\mathbf{z_1}$, and we estimate the sample at time $t$ by using a neural network to compute $\mathbf{\hat{x_t}} \sim f_{\theta}(\mathbf{z_t}; t)$. Then, we perform the same forward process again at a lower noise level $r$ where $r < t$ such that $\mathbf{z_r} = \alpha_r \mathbf{\hat{x}} + \sigma_r \mathbf{\hat{\epsilon}}$, where $\hat{\boldsymbol{\epsilon}} = (\mathbf{z_t} - \alpha_t \hat{\mathbf{x}}) / \sigma_t$. After repeatedly executing this process of predicting the clean sample and performing the forward process at a lower noise level, we finally arrive at the clean sample. In this algorithm, the randomness is only induced from the initial sample $\mathbf{z_1}$, and the rest of the reverse sampling process is deterministic. In practice, this is the DDIM sampling technique~\cite{song2022denoisingdiffusionimplicitmodels}, which is used for efficient sampling from diffusion models without performing the reverse process across the entire noise schedule.

We can also formulate the forward corruption process of diffusion models with a stochastic differential equation (SDE): $$d\mathbf{z_t} = f_t\mathbf{z_t}dt + g_td\mathbf{z},$$
where $d\mathbf{z}$ is a Brownian motion that characterizes an infinitesimal Gaussian, and $(f_t, g_t)$ are the parameters of the noise schedule~\cite{gao2025diffusionmeetsflow,karras2022elucidatingdesignspacediffusionbased,song2021scorebasedgenerativemodelingstochastic}. To sample from the model, we perform the reverse process using the score $\nabla \log p_t$ of the forward process as the following: $$d\mathbf{z_t} = \left(f_t\mathbf{z_t} - \frac{1 + \eta_t^2}{2}g_t^2 \nabla \log p_t(\mathbf{z_t})\right)dt + \eta_t g_t d\mathbf{z}.$$ This formulation contains an additional parameter $\eta_t$ which controls the level of stochasticity at inference time. In practice, this term is also related to the ``churn,'' which is the fraction of the reverse step to undo by renoising~\cite{karras2022elucidatingdesignspacediffusionbased}.

\subsection{Flow Matching}
Flow matching models learn a transformation function that maps data from a simple base distribution to the complex target data distribution by matching the flow of probability densities~\cite{genmatching,lipman2023flowmatchinggenerativemodeling}. Flow matching views the forward noising process as a linear interpolation between the data $\mathbf{x}$ and noise $\boldsymbol{\epsilon}$: $$\mathbf{z_t} = t \boldsymbol{\epsilon} + (1 - t)\mathbf{x}.$$ We can see that if we set $\alpha_t=1-t$ and $\sigma_t=t$, then this formulation is equivalent to the Gaussian diffusion forward process~\cite{albergo2023stochasticinterpolantsunifyingframework,gao2025diffusionmeetsflow}.

Assume that our initial sample $\mathbf{z_1}$ is pure Gaussian noise. If we let $r < t$, then we can derive $\mathbf{z_t} = \mathbf{z_r} + (t - r)\mathbf{v}$, where $\mathbf{v} = \boldsymbol{\epsilon} - \mathbf{x}$, which is the ``flow'' or ``vector field'' learned by a neural network $f_\theta$. We can sample $\mathbf{z_r}$ from $\mathbf{z_t}$ by simply going reverse in time and estimating the flow at time $t$ as $\hat{\mathbf{v}} = f_\theta(\mathbf{z_t}; t) = \hat{\boldsymbol{\epsilon}} - \hat{\mathbf{x}}$. We then compute $\mathbf{z_r}$ as $$\mathbf{z_r} = \mathbf{z_t} + (r - t)\hat{\mathbf{v}},$$ and we repeatedly perform this reverse process starting with $\mathbf{z_1}$ till we reach the clean sample.

We can also formulate flow matching with an ordinary differential equation (ODE)~\cite{lipman2023flowmatchinggenerativemodeling}: $$d\mathbf{z_t} = \mathbf{u_t}dt,$$ where $\mathbf{z_t} = \alpha_t \mathbf{x} + \sigma_t \boldsymbol{\epsilon}$ and $\mathbf{u_t} = \dot{\alpha_t} \mathbf{x} + \dot{\sigma_t} \boldsymbol{\epsilon}$. To generate from the model, we simply reverse the ODE in time. However, in practice, we typically also include a score-correction term since we are approximating $\mathbf{u_t}$. Formally, the reverse-time probability flow ODE to generate samples is: 
$$d\mathbf{z_t} = \left(\mathbf{u_t} - \frac{1}{2}\epsilon_t^2 \nabla \log p_t(\mathbf{z_t})\right)dt,$$
where $\epsilon_t^2$ is now a deterministic scaling term. We can also write it as an SDE under a special case of stochastic interpolants~\cite{albergo2023stochasticinterpolantsunifyingframework,albergo2023buildingnormalizingflowsstochastic}:
$$d\mathbf{z_t} = \left(\mathbf{u_t} - \frac{1}{2}\epsilon_t^2 \nabla \log p_t(\mathbf{z_t})\right)dt + \epsilon_t d\mathbf{z},$$ where in this formulation, $\epsilon_t$ can be tuned to control the level of stochasticity during generation. 

\subsection{Equivalence of Diffusion and Flow Matching}
One way to show equivalence between diffusion and flow matching is by deriving the parameters of one generative Markov process from the other. This is simpler to derive using the SDE / ODE perspective of both diffusion / flow matching~\cite{albergo2023stochasticinterpolantsunifyingframework,albergo2023buildingnormalizingflowsstochastic,gao2025diffusionmeetsflow}. 

In diffusion, we found an SDE parameterized by $f_t, g_t, \eta_t$. Consider the deterministic evolution that would occur if we only followed the drift $f_t\mathbf{z_t}$ without noise. If we start with $\mathbf{z_0} = \mathbf{x}$ and have it evolve under the ODE $\mathbf{\dot{z}_t} = f_t\mathbf{z_t}$ then $\mathbf{z_t} = \alpha_t \mathbf{x}$. Thus, $\alpha_t$ encodes how a deterministic initial condition would scale over time, so we can formulate it as: $$\alpha_t = \exp\left(\int_0^t f_r dr\right).$$
Next, we consider how noise accumulates relative to this deterministic scaling. The noise intensity $g_t$ is ``modulated''  by the factor $\exp \left(-2 \int_0^r f_u du \right)$ to account for how the drift $f_t$ stretches or shrinks the space. This leads to $$\sigma_t = \left(\int_0^t g^2_r \exp\left(-2 \int_0^r f_u d_u\right) dr \right)^{1/2}.$$
Since we factor out the deterministic scaling from the SDE, the remaining effective noise standard deviation after evolving from 0 to $t$ is precisely this expression.
Finally, the level of stochasticity in the flow matching framework, $\mathbf{\epsilon_t}$, must match the scaled version of the diffusion noise. Thus, we have $$\mathbf{\epsilon_t} = \eta_t g_t.$$
We leave the derivation from flow matching to diffusion as an exercise for the reader.

\section{Generator Matching}

\subsection{Marginal Paths}

Generator Matching builds on the previous processes by generalizing the framework that they operate under. \cite{genmatching, fairmatching} As referenced in the previous section, what Diffusion and flow matching do, in essence, is define a marginal probability path $(p_t(dx))_{0 \leq t \leq 1}$, a sequence of probability distributions along which samples from some starting (noise) distribution at time 0, $p_{simple} = p_0$ are transformed into some valid sample from the data distribution at time 1, $p_{data} = p_1$. The first step towards solving for the marginal probability path is choosing the prior $p_{sample}$ and the conditional path $p_t(dx|z)$. Common choices for the conditional path are mixtures with tunable parameter $\kappa$: 

$$
    p_t(dx | z) = (1 - \kappa_t) \cdot p_0(dx) + \kappa_t \cdot \delta_z(dx)
$$

Or for state spaces $S=\mathbb{R}^d$, the geometric average with $\delta_z$ as the delta distribution: 
$$p_t(dx | z) = \mathbb{E}_{x_0} \left[ \delta_{\sigma_t x_0 + \alpha_t z}(dx) \right]$$

\subsection{Markov Processes and Generators}

Markov processes can be fully defined by their transition kernel, $(k_{t+h|t})_{0 \leq t < t + h \leq 1}$, which defines for all $x$ $k_{t+h|t}(\cdot|x)$ where  $\mathbb{P}[X_{t+h} \in A \mid X_t = x]$. \cite{genmatching} With Generator Matching, we use the concept of finding the marginal probability path to model a Markov process with $X_0 \sim p_0$ and simulating $X_{t+h} \sim k_{t+h|t}(\cdot|X_t)$. Thus, if we can approximate the Markov kernel, then we can approximate the entire Markov process and by extension the marginal probability path.

A first order Taylor approximation of the transition kernel with error term $o(h)$ can be written as: 
$$k_{t+h|t} = k_{t|t} + h\mathcal{L}_t + o(h), \quad
\mathcal{L}_t := \left. \frac{d}{dh} \right|_{h=0} k_{t+h|t}, \quad
k_{t|t}(\cdot|x) = \delta_x$$
$\mathcal{L}_t$, the first derivative, is called the generator of the transition kernel. 

However, this representation is only defined when $p_t$ has a defined density. In order to allow a tractable comparison of probability distributions $p_t$, we also introduce a set of test functions $\mathcal{T}$ such that for each function $f \in \mathcal{T}$, the values 

$$\langle p_t, f\rangle := \int f(x) p_t(dx) = \mathbb{E}_{x \sim p_t}\left[f(x)\right]$$ 

fully characterize the distribution. 
Furthermore, we can generalize this notion to express generators of the transition kernel, given by 

$$\left\langle k_{t+h \mid t}, f\right\rangle(x) \:= \left\langle k_{t+h \mid t}(\cdot \mid x), f\right\rangle=\mathbb{E}\left[f\left(X_{t+h}\right) \mid X_t=x\right]$$

This allows us to associate the generators with the differentiable functions 
$$\left.\frac{d}{d h}\right|_{h=0}\left\langle k_{t+h \mid t}, f\right\rangle(x)=\lim _{h \rightarrow 0} \frac{\left\langle k_{t+h \mid t}, f\right\rangle(x)-f(x)}{h} \stackrel{\text { def }}{=}\left[\mathcal{L}_t f\right](x)$$

Under regularity assumptions, there is a 1:1 correspondence between Markov processes and their generators. For certain spaces, particularly those that commonly appear in deep learning applications, generators can be fully defined. For a space $S$ that is discrete and has $|S| < \infty$, such as languages for use in language generation, the generator $\mathcal{L}_t$ is given by the rate transition matrix $Q_t$, and thus the corresponding Markov process is a CTMC. 

Likewise, for Euclidean spaces $S=\mathbb{R}$, $$\mathcal{L}_t f(x) = \nabla f(x)^T u_t(x) + \frac{1}{2} \nabla^2 f(x) \cdot \sigma_t^2(x) + \int \left[f(y) - f(x)\right] Q_t(dy; x)$$ 

The first term corresponds to flow processes, where $u$ is a velocity field, the second term corresponds to diffusion, where $\sigma$ is a diffusion coefficient, and the third term corresponds to "jump" where $Q_t$ is a finite measure. By learning these parameters with a neural network, we can parameterize the process and derive its generator.

\subsection{Finding the Marginal Generator Using the Kolmogorov Forward Equation}

In addition to allowing for a neural net to parameterize the Markov process, we can also check to see if that Markov process produces an intended marginal probability path. The generator determines these marginal probabilities, as governed by the following relationship, known as the Kolmogorov Forward Equation (KFE) \cite{genmatching}: 

$$\partial \langle p_t, f \rangle = \langle p_t, \mathcal{L}_tf\rangle$$

This means that marginal probabilities of a generator can be recovered if that generator is specified. The converse also holds true, namely that given some marginal probability path, we can recover the generator that satisfies the KFE. 

With this stated, we can connect the original construction of conditional probability paths to our marginal generator. If we calculate conditional generators for the conditional probability path, then by the equation $\mathcal{L}_t f(x) = \mathbb{E}_{z \sim p_{1|t}(\cdot | x)} \big[ \mathcal{L}_t^z f(x) \big]$ we can calculate the final marginal generator from the originally specified conditional paths in 3.1: 

$$\mathcal{L}_tf(x) = \nabla f(x)^T \mathbb{E}_{z \sim p_{1|t}(\cdot | x)} \big[ u_t(x | z) \big] 
+ \frac{\nabla^2 f(x)}{2} \cdot \mathbb{E}_{z \sim p_{1|t}(\cdot | x)} \big[ \sigma_t^2(x | z) \big] 
+ \int \big[ f(y) - f(x) \big] \mathbb{E}_{z \sim p_{1|t}(\cdot | x)} \big[ Q_t(dy; x | z) \big]
$$

Where $u_t(x | z)$, $\sigma_t^2(x | z)$, and $Q_t(dy; x | z)$ are all conditional generators.

\subsection{Training the Parameterized Generator}

Lastly, to derive the final generator, it remains to train a parameterized version of it to approximate the true underlying Markov generator. To this end, if we take the generator $\mathcal{L}_t f(x) = \mathbb{E}_{z \sim p_{Z|t}(\cdot | x)} \big[ \mathcal{L}_t^z f(x) \big]
$ linearly parameterized by $F_t(x) = \mathbb{E}_{z \sim p_{Z|t}(\cdot | x)} \big[ F_t(x|Z) \big]
$ then the Generator Matching loss is given by $$\mathcal{L}_{\text{GM}}(\theta) = \mathbb{E}_{t\sim \text{Unif}, x \sim p_t} D_{X_t} \big( F_t(x), F_t^\theta(x) \big)
$$

Where $F_t^\theta(x)$ is the neural network and D is the Bregman Divergence, $D(a, b) = \phi(a) - \big[ \phi(b) + \langle a - b, \nabla \phi(b) \rangle \big]
$ for strictly convex function $\phi$.

Of course, we don't know the marginal generator or a linear parameterization of it, so to make this problem tractable we linearize the conditional generator $\mathcal{L}_{\text{CGM}}(\theta) = \mathbb{E}_{t\sim \text{Unif}, z \sim p_{\text{data}},x \sim p_t(\cdot | z)} D_{X_t} \big( F_t^z(x), F_t^\theta(x) \big)
$. 

Since these losses have the same gradients, minimizing these objectives is the same, and we can simply assume that $F_t(x)$ has the shape $F_t(x) = \int F_t^z(x)p_{1|t}(dz|x)$.

Thus, there is now a tractable and scalable way to train Generator Matching models that allow for broad characterization and approximation of Markov generators while also having the benefits detailed below in Section~\ref{section::comparison}.

\subsection{Markov Superposition with Generator Matching}

Given generators that satisfy Markov processes as specified by the above Generator Matching process, we can add generators together without violating the properties of a Generator Matching model. The reason this holds true is because a generator $\mathcal{L}_t$ is a linear operator and the KFE is a linear equation, so for coefficients $a, b \geq 0$ where $a + b = 1$, and generators $\mathcal{L}_t$ and $\mathcal{L}_t^\prime$ satisfying the KFE, 

$$\langle p_t, \left(a \mathcal{L}_t + b \mathcal{L}^\prime_t \right)f\rangle = a \langle p_t, \mathcal{L}_t f\rangle + b \langle p_t, \mathcal{L}_t^\prime f\rangle = a \partial \langle p_t, f \rangle + b \partial \langle p_t, f \rangle = \partial \langle p_t, f \rangle$$
so the superposition $a \mathcal{L}_t + b \mathcal{L}_t$ must also satisfy the KFE.
From this, it directly follows that we can directly combine parameterized models built under the generating matching framework, as a result enabling combination of diffusion, flow, and jump processes within a single Markov generator. This results in a more robust model, leading to both a greater diversity in generated data and closer approximations to the underlying generator, as shown in \cite{genmatching}. 

\section{Diffusion vs. Flow Matching Under Generator Matching}
\label{section::comparison}
\subsection{Unification with Generator Matching}
Within the Generator Matching framework, both diffusion-based and flow-matching models can be understood as constructing Markov processes that connect a simple prior distribution $p_0$ to a target distribution $p_1$. Each model corresponds to solving a KFE or probability flow PDE to evolve distributions over time~\cite{genmatching,lipman2023flowmatchinggenerativemodeling,song2021scorebasedgenerativemodelingstochastic}. Both diffusion and flow matching are ultimately satisfy the same fundamental continuity equations. However, the presence or absence of a stochastic component leads to different forms of these equations (stochastic SDE-based KFE for diffusion vs. deterministic PDE for flow). This allows us to compare the robustness of both diffusion and flow matching under a single theoretical framework.

\subsection{Comparison Under Generator Matching}
For diffusion-based generative models, the underlying PDE is second-order~\cite{song2021scorebasedgenerativemodelingstochastic}. Specifically, it includes a diffusion term characterized by a Laplacian operator. The forward diffusion process takes $p_0$ and produces a smoothed final distribution $p_1$ at time $t=1$. While this forward direction is stable and well-defined, the inverse operation---recovering $p_0$ from $p_1$ by running the process backward---is fundamentally ill-posed. Inverting a second-order parabolic PDE is known to be highly sensitive to perturbations: small errors in the estimated drift or density can lead to disproportionately large deviations in the reconstructed initial distribution~\cite{vazquez2024backsteppingpartialdifferentialequations}. Moreover, because diffusion smooths and aggregates paths of the process, distinct initial states may map to nearly identical final distributions, destroying uniqueness and making the backward solution non-unique and prone to instability. Thus, the generative reverse-time SDE for sampling in diffusion models is less robust and more sensitive to model imperfections.

In contrast, flow matching models are governed by first-order PDEs corresponding to optimal transport or deterministic drift processes~\cite{lipman2023flowmatchinggenerativemodeling}. The generator of such a process is a first-order operator that shifts probability mass along trajectories defined by a velocity field. The solution of a first-order PDE less susceptible to small errors in the velocity field, which would not lead to exponential blow-up or severe instability~\cite{albergo2023buildingnormalizingflowsstochastic}. Since no smoothing operation is imposed, the mapping between the initial and final distributions remains more directly invertible, ensuring that small approximation inaccuracies do not severely compromise the generative process. This means that flow matching provides a more stable and robust approach under the Generator Matching framework, as it avoids the intrinsic difficulties associated with inverting a smoothing (second-order) diffusion operator.

The Generator Matching viewpoint clarifies that the relative robustness of flow matching stems from its reliance on first-order PDEs, which naturally allow increased stability in the forward and backward evolution of distributions, in contrast to the sensitive backward inversion of a second-order diffusion process.

\subsection{Mixing Diffusion and Flow Matching}
From the Generator Matching perspective, the boundary between purely diffusive and purely deterministic flow-based generative models is not fixed. Instead, there is a continuum of possible intermediates that blend stochastic and deterministic components~\cite{genmatching}. For instance, consider a “denoising diffusion” model: it can be viewed as a flow model trained via the CGM objective with a mean squared error loss. In this setting, sampling can be made partially stochastic by incorporating a divergence-free, Langevin-like dynamics component on top of the deterministic flow field. This flexible framework opens the door to new modeling avenues. Traditional diffusion models typically fix the diffusion coefficient, often resulting in a uniform noise schedule that is not tailored to the geometry of the underlying data manifold. Under the Generator Matching lens, one could instead learn a state-dependent diffusion coefficient $\sigma_t(x)$, selecting how much noise to inject dynamically based on the current distribution and local structure of the manifold~\cite{genmatching}.

One potential new method to combine diffusion and flow matching models is using a learned, state-dependent noise schedule $\sigma_t(x)$~\cite{lee2024antadaptivenoiseschedule}. We can use such a schedule to create a generative process that adaptively toggles between diffusion-like smoothing and flow-like deterministic transport as needed. In regions of the data manifold where the distribution is complex or highly curved, introducing more noise can regularize the generation process, prevent mode collapse, and facilitate better coverage of the data's support. Conversely, in well-understood or relatively flat regions, the model could rely more heavily on deterministic flow, minimizing unnecessary stochasticity and focusing on precise, efficient mass transport. This adaptive strategy could dynamically balance the strengths of both paradigms—stability and smoothing from diffusion, and controlled, invertible transport from flows—leading to generative models that are both more robust and more data-efficient.

\section{Conclusion}
Through the lens of Generator Matching, we show how to unify and compare diffusion and flow matching models under a robust theoretical framework. Our analysis provides a more nuanced theoretical understanding on why flow matching models are more robust than diffusion models under specific constraints, and how we can apply Generator Matching to benefit from both paradigms. We hope to inspire future research in creating novel generative Markov processes for underexplored data modalities with strong theoretical foundations.

\newpage
{
\bibliographystyle{plain}
\bibliography{biblio}
}
\end{document}